\documentclass{article}
\usepackage{spconf,amsmath,graphicx,hyperref}
\usepackage{booktabs}   
\usepackage{multirow}   
\usepackage{makecell}   
\usepackage[table]{xcolor} 
\usepackage{graphicx}
\usepackage{amssymb}
\usepackage{pifont}
\usepackage{siunitx} 
\usepackage{cleveref}

\title{PhaseMark: A Post-hoc, Optimization-Free Watermarking of AI-generated Images in the Latent Frequency Domain}
\name{Sung Ju Lee \qquad Nam Ik Cho
\thanks{}
}
\address{Department of ECE \& INMC, Seoul National University, Korea}

\usepackage{eso-pic}
\AddToShipoutPictureBG*{%
    \AtPageUpperLeft{%
        \put(\LenToUnit{2cm},\LenToUnit{-27cm}){
            \parbox{\textwidth}{%
                \footnotesize
                \copyright 2026 IEEE. Personal use of this material is permitted. Permission from IEEE must be obtained for all other uses, in any current or future media, including reprinting/republishing this material for advertising or promotional purposes, creating new collective works, for resale or redistribution to servers or lists, or reuse of any copyrighted component of this work in other works.
            }%
        }%
    }%
}

\begin{document}
%
\maketitle
\begin{abstract}
The proliferation of hyper-realistic images from Latent Diffusion Models (LDMs) demands robust watermarking, yet existing post-hoc methods are prohibitively slow due to iterative optimization or inversion processes. We introduce PhaseMark, a single-shot, optimization-free framework that directly modulates the phase in the VAE latent frequency domain. This approach makes PhaseMark thousands of times faster than optimization-based techniques while achieving state-of-the-art resilience against severe attacks, including regeneration, without degrading image quality. We analyze four modulation variants, revealing a clear performance-quality trade-off. PhaseMark demonstrates a new paradigm where efficient, resilient watermarking is achieved by exploiting intrinsic latent properties.
\end{abstract}
\begin{keywords}
Latent Phase, Watermarking, LDMs
\end{keywords}

\section{Introduction}
\label{sec:intro}
Recent advances in Latent Diffusion Models (LDMs)~\cite{rombach2022high} enable hyper-realistic image generation, necessitating robust watermarking for copyright protection and content authentication. However, traditional frequency-based methods~\cite{cox2007digital, nikolaidis1996copyright} are vulnerable to modern threats like regeneration attacks, requiring a new paradigm for the generative AI era.

Recent post-hoc latent space watermarking methods show resilience to regeneration attacks but suffer from prohibitive computational costs. For instance, ZoDiac~\cite{zhang2025attack} and FreqMark~\cite{guo2024freqmark} rely on a multi-minute, per-image optimization process; other prominent methods employ DDIM inversion~\cite{zhang2025attack, wen2024tree, ci2024ringid, yang2024gaussian, lee2025semantic}, which requires multiple expensive LDM executions for detection. This high computational barrier impedes real-world deployment.

We propose PhaseMark, a single-shot, optimization-free post-hoc watermarking framework that overcomes these limitations. PhaseMark directly modulates the phase in the mid-band frequency of the VAE latent vector. Our contributions are: First, PhaseMark is thousands of times faster than optimization-based methods while achieving comparable or superior resilience to regeneration attacks. Second, we design and analyze four variants along the axes of \textit{absolute vs. relative phase} and \textit{hard vs. soft modulation}, revealing a clear trade-off between performance and image quality. We demonstrate that expensive optimization is not a prerequisite for resilience; rather, it is achievable by efficiently exploiting the latent space's intrinsic properties.

\section{Latent Phase Watermarking}
\label{sec:methods}
PhaseMark is a fast, robust, post-hoc watermarking model that directly modulates the phase in a VAE's latent frequency domain in a single-shot, achieving high efficiency and robustness without iterative optimization.

\subsection{Overall Framework}
\label{ssec:methods-framework}
The VAE encoder first maps an input image $\mathcal{I}$ to a latent vector $z_0$. For spatial robustness, we apply a 2D FFT to its central region, $z_{crop}$, yielding $\mathcal{F}(z_{crop})$. An $L$-bit message $M$ is embedded into $L$ $2 \times 2$ blocks, $\mathcal{B}= \{ B_1, \ldots, B_L \}$, extracted from stable mid-band frequencies. 
This selection avoids low-frequency image content and high-frequency noise. Block selection uses axis offsets to bypass principal axes. 
After modulation, Hermitian symmetry is enforced to ensure a real-valued output after the Inverse FFT. Finally, the VAE decoder reconstructs the watermarked image $\mathcal{I}'$.

\subsection{Phase Modulation for Embedding}
\label{ssec:methods-embedding}
To embed a message bit $m_i$ into each extracted block $B_i$, we propose four phase modulation strategies, which are categorized into two distinct families.

\subsubsection{Absolute Phase Modulation}
\label{sssec:methods-AbsolutePM}
Methods in this family modulate the phase of complex numbers within a block to predefined absolute values, which are independent of other elements.

\noindent
\textbf{Absolute Phase Modulation (APM).}
APM, a \textit{hard} modulation method, forcibly assigns the phase of each complex number $c$ in block $B_i$ based on the message bit $m_i$:
\begin{equation}
\label{eq:apm}
\angle c' = (-1)^{1-m_i} \frac{\pi}{2}
\end{equation}

\noindent
\textbf{Phase Constellation Quantization (PCQ).}
This \textit{soft} modulation method is designed to minimize image distortion. Two phase constellation sets $\mathcal{P}_0$ and $\mathcal{P}_1$ are predefined based on the bit value. The original phase $\angle c$ of each complex number $c$ within block $B_i$ is then quantized to the nearest phase in the constellation set corresponding to the bit:
\begin{equation}
\label{eq:pcq}
\angle c' = 
\underset{p \in \mathcal{P}_{m_i}}{\arg\min} |\angle c - p |_{\text{angle}}
\end{equation}
Here, $|\cdot|_{\text{angle}}$ denotes the minimum angular distance.

\subsubsection{Relative Phase Modulation}
\label{sssec:methods-RelativePM}
This family of methods encodes information in the relative phase relationships among elements within a block. For a $2 \times 2$ block $B_i = \{c_1, c_2, c_3, c_4 \}$ the elements $c_1$ and $c_3$ serve as anchors.

\noindent
\textbf{Intra-block Phase Synchronization (IPS).}
A \textit{hard} relative modulation method that forcibly synchronizes the phases of $c_2$ and $c_4$ with those of the anchors $c_1$ and $c_3$, while preserving their original magnitudes.
\begin{equation}
\label{eq:ips}
\begin{aligned}
    \angle c_2' &\leftarrow \angle c_1 + (1-m_i)\pi \\
    \angle c_4' &\leftarrow \angle c_3 + (1-m_i)\pi
\end{aligned}
\end{equation}

\noindent
\textbf{Soft Phase Synchronization (SPS).}
SPS, a \textit{soft} method, interpolates vectors $c_2$ and $c_4$ towards their phase-synchronized targets. The modified vector $c_k'$ for $k \in \{ 2, 4 \}$
is a linear interpolation between the original $c_k$ and its target $c_{k,tgt.}$:
\begin{equation}
\label{eq:sps}
\begin{aligned}
c_k' &= (1-\gamma)c_k + \gamma \cdot c_{k, \text{tgt.}} \quad \text{for } k \in \{ 2, 4 \},
\end{aligned}
\end{equation}
where $c_{k, \text{tgt.}} = |c_k|e^{j\phi_{k, \text{tgt.}}}$ and $\phi_{k, \text{tgt.}} = \angle c_{k-1} + (1-m_i)\pi$.
Here, $\gamma \in [0,1]$ is an interpolation strength parameter. Note that when $\gamma = 1$, SPS becomes identical to IPS.

\begin{table*}[th]
\centering
\scriptsize 
\caption{Comparison of detection performance (TPR@1\%FPR) and image quality ($\Delta$FID-1k) against baseline methods under various attacks. Ours-Perf (APM) and Ours-Qual (PCQ) denote our performance- and quality-centric models, respectively.}
\begin{tabular}{@{}c@{\hspace{2mm}} l @{\hspace{2mm}} c @{\hspace{2mm}} c| ccccc ccc cc c S[table-format=1.3, table-space-text-post={**}] }
\arrayrulecolor{black}
& & & \multicolumn{2}{c}{} & \multicolumn{4}{c}{Signal Processing Attacks} & \multicolumn{3}{c}{Regeneration Attacks} & \multicolumn{2}{c}{Cropping Attacks} & & \textit{Quality} \\
\cmidrule[\heavyrulewidth](r){2-15}
\cmidrule[\heavyrulewidth](lr){16-16}
& Methods & Post-hoc &  Bits & Clean & \multicolumn{1}{|c}{Cont.} & JPEG & Blur & BM3D & \multicolumn{1}{|c}{VAE-B} & VAE-C & Diff. & \multicolumn{1}{|c}{C.C.} & R.C. & \multicolumn{1}{|c}{Avg} & {$\Delta$FID-1k$\downarrow$} \\
\cmidrule(r){2-15}
\cmidrule(lr){16-16}
\multirow{11}{*}{\rotatebox{90}{Verification Task}}
& DwtDctSvd & \checkmark & 32 & 1.000 & 0.106 & 0.173 & 1.000 & 0.609 & 0.205 & 0.060 & 0.555 & 1.000 & 0.229 & 0.494 & -0.389 \\
& RivaGAN & \checkmark & 32 & 1.000 & 0.997 & 0.798 & 1.000 & 0.972 & 0.025 & 0.016 & 0.998 & 0.998 & 0.119 & 0.692 & -0.785 \\
& S.Sign. & \ding{55} & 48 & 1.000 & 0.993 & 0.927 & 0.990 & 0.933 & 0.677 & 0.667 & 0.997 & 0.999 & 0.003 & 0.819 & -0.241 \\
& G.Shading & \ding{55} & 256 & 1.000 & 1.000 & 1.000 & 1.000 & 1.000 & 0.996 & 1.000 & 1.000 & 1.000 & 1.000 & 1.000 & -0.317 \\
\cmidrule(r){2-15}
\cmidrule(lr){16-16}
& Tree-Ring & \ding{55} & zero-bit & 0.957 & 0.900 & 0.548 & 0.934 & 0.815 & 0.509 & 0.536 & 0.543 & 0.509 & 0.734 & 0.699 & 1.184 \\
& RingID & \ding{55} & zero-bit & 1.000 & 1.000 & 1.000 & 1.000 & 1.000 & 0.992 & 1.000 & 1.000 & 1.000 & 1.000 & 0.999 & 2.054 \\
& ZoDiac & \checkmark & zero-bit & 0.998 & 0.998 & 0.973 & 0.998 & 0.997 & 0.944 & 0.958 & 0.972 & 0.989 & 0.995 & 0.982 & 0.091 \\
& SFW-HSQR & \ding{55} & zero-bit & 1.000 & 1.000 & 1.000 & 1.000 & 1.000 & 0.992 & 1.000 & 1.000 & 1.000 & 1.000 & 0.999 & -0.721 \\
\cmidrule(r){2-15}
\cmidrule(lr){16-16}
& Ours-Qual. (PCQ) & \checkmark & 128 & 1.000 & 0.988 & 0.870 & 0.998 & 0.992 & 0.918 & 0.933 & 1.000 & 0.999 & 0.998 & 0.970 & -0.517 \\
& Ours-Perf. (APM) & \checkmark & 128 & 1.000 & 1.000 & 1.000 & 1.000 & 1.000 & 0.996 & 0.997 & 1.000 & 1.000 & 1.000 & 0.999 & 0.042 \\
\cmidrule[\heavyrulewidth](r){2-15}
\cmidrule[\heavyrulewidth](lr){16-16}

\\ [-3mm]

\cmidrule[\heavyrulewidth](r){2-15}
& Methods & Post-hoc & User IDs & Clean & \multicolumn{1}{|c}{Cont.} & JPEG & Blur & BM3D & \multicolumn{1}{|c}{VAE-B} & VAE-C & Diff. & \multicolumn{1}{|c}{C.C.} & R.C. & \multicolumn{1}{|c}{Avg} \\
\cmidrule(r){2-15}
\multirow{11}{*}{\rotatebox{90}{Identification Task}}
& DwtDctSvd & \checkmark & $10^6$ & 1.000 & 0.019 & 0.000 & 0.999 & 0.037 & 0.000 & 0.000 & 0.000 & 0.000 & 0.003 & 0.206 \\
& RivaGAN & \checkmark & $10^6$ & 0.974 & 0.772 & 0.023 & 0.961 & 0.348 & 0.000 & 0.000 & 0.852 & 0.909 & 0.000 & 0.484 \\
& S.Sign. & \ding{55} & $10^6$ & 0.989 & 0.948 & 0.067 & 0.632 & 0.285 & 0.008 & 0.008 & 0.970 & 0.982 & 0.000 & 0.489 \\
& G.Shading & \ding{55} & $10^6$ & 1.000 & 1.000 & 0.999 & 1.000 & 1.000 & 0.992 & 0.999 & 1.000 & 1.000 & 1.000 & 0.999 \\
\cmidrule(r){2-15}
& Tree-Ring & \ding{55} & 2048 & 0.303 & 0.207 & 0.072 & 0.256 & 0.162 & 0.083 & 0.072 & 0.054 & 0.009 & 0.033 & 0.125 \\
& RingID & \ding{55} & 2048 & 1.000 & 1.000 & 0.975 & 1.000 & 0.996 & 0.978 & 0.970 & 0.998 & 0.874 & 0.978 & 0.977 \\
& ZoDiac & \checkmark & 64 & 0.164 & 0.038 & 0.000 & 0.080 & 0.027 & 0.001 & 0.002 & 0.000 & 0.000 & 0.000 & 0.031 \\
& SFW-HSQR & \ding{55} & 8192 & 1.000 & 1.000 & 0.990 & 1.000 & 0.999 & 0.976 & 0.982 & 1.000 & 1.000 & 0.999 & 0.995 \\
\cmidrule(r){2-15}
& Ours-Qual. (PCQ) & \checkmark & $10^6$ & 1.000 & 0.848 & 0.427 & 0.972 & 0.849 & 0.573 & 0.592 & 0.933 & 0.981 & 0.989 & 0.816 \\
& Ours-Perf. (APM) & \checkmark & $10^6$ & 1.000 & 1.000 & 0.984 & 1.000 & 0.999 & 0.945 & 0.954 & 1.000 & 1.000 & 1.000 & 0.988 \\
\cmidrule[\heavyrulewidth](r){2-15}
\end{tabular}
\label{tab:exp_detection}
\end{table*}

\subsection{Watermark Detection}
\label{ssec:methods-detection}
Detection reverses the embedding process to extract blocks $\mathcal{B}'$ and message $\hat{m}$. The bit $\hat{m}_i$ from each block $B_i'$ is determined as follows:

\noindent
\textbf{APM.} The sign of the sum of all phases within the block determines the bit. If $\Sigma_{c \in B_i'} \angle c > 0$, then $\hat{m}_i = 1$; otherwise, $\hat{m}_i = 0$.

\noindent
\textbf{PCQ.} The bit is determined by comparing the total distance from the block's phases to each of the two constellation sets, $\mathcal{P}_0$ and $\mathcal{P}_1$.

\noindent
\textbf{IPS \& SPS.} A correlation score which represents the phase relationship with the anchor points is calculated: $S_i = \cos(\angle c_1 - \angle c_2) + \cos(\angle c_3 - \angle c_4).$ If $S_i > 0$, then $\hat{m}_i = 1$; otherwise, $\hat{m}_i = 0$.

A watermark is detected if the Bit Accuracy (BA) exceeds a threshold $\tau$ set for a 1\% False Positive Rate (FPR). The verification ($\tau_{vrf.}$) and identification ($\tau_{idf.}$) thresholds are defined as:
$\tau_{vrf.} = \frac{1}{L} \min \{ k \mid P(\text{matches} \ge k) \le \alpha \}$.
For identification among $N = 10^6$ users, a more stringent threshold is derived using a Bonferroni correction: 
$\tau_{idf.} = \frac{1}{L} \min \{ k \mid P(\text{matches} \ge k) \le \alpha / N \}$.
The performance for both tasks is evaluated as the True Positive Rate (TPR) at the prespecified 1\% FPR.

\begin{table}[th]
\centering
\scriptsize 
\caption{Computational efficiency comparison. PhaseMark's embedding and detection times are orders of magnitude faster than ZoDiac, enabling real-time processing.}
\begin{tabular}{@{}l @{\hspace{2mm}} c @{\hspace{0.5mm}} c @{\hspace{1.3mm}} c @{\hspace{1.5mm}} | @{\hspace{1.5mm}} c @{\hspace{2mm}} c@{}}
\arrayrulecolor{black}
\toprule
Methods & Approach & Opt.Free & Train Cost & Latency-${\mathcal{E}}$ & Latency-${\mathcal{D}}$ \\
\midrule
RivaGAN & End-to-end Training & \ding{55} & 300 epochs & 0.416 s & 0.399 s \\
S.Sign. & End-to-end Training & \ding{55} & $\geq$ 50 epochs & - & 0.101 s \\
\midrule
ZoDiac & Iterative Optimization & \ding{55} & 100 iters/img & 7.334 m & \multirow{2}{*}{4.113 s} \\
\textit{Semantic} & Iterative Inversion (${\mathcal{D}}$) & \checkmark & - & - & \\
\midrule
Ours & Single-Shot Transform & \checkmark & - & 0.142 s & 0.050 s \\
\bottomrule
\end{tabular}
\label{tab:exp_computation}
\end{table}

\section{Experiments}
\label{sec:experiments}
\subsection{Experimental Setup and Implementation Details}
\label{ssec:exp-setup}
\noindent
\textbf{Datasets \& Generation.}
We generate 1,000 $512 \times 512$ images per dataset using Stable Diffusion v2-1-base model~\cite{rombach2022high} ($\text{CFG}=7.5$, 50 DDIM steps). Prompts are from MS-COCO-2017~\cite{lin2014microsoft}, with SD-Prompts~\cite{gustavo} and DiffusionDB~\cite{wang2022diffusiondb} for generalization.

\noindent
\textbf{Watermarking Setup.}
We use the LDM's VAE. Latent crop $z_{crop}$ is $44 \times 44 \times 4$, mid-band frequencies are radially bounded by $[10,18]$, and the message is 32 bits per channel.
We set $\gamma = 0.8$ for SPS.

\noindent
\textbf{Baselines.}
We compare against post-hoc (DwtDctSvd~\cite{cox2007digital}, RivaGAN~\cite{zhang2019robust}, ZoDiac~\cite{zhang2025attack}) and in-generation (S.Sign.~\cite{fernandez2023stable}, Tree-Ring~\cite{wen2024tree}, RingID~\cite{ci2024ringid}, G.Shading~\cite{yang2024gaussian}, SFW~\cite{lee2025semantic}) methods.

\noindent
\textbf{Attacks.}
We evaluate robustness against nine attacks: contrast (0.5), JPEG ($\text{Q}=25$), Gaussian blur ($r=5$), BM3D ($\sigma=0.1$), VAE regeneration (VAE-B~\cite{balle2018variational} and VAE-C~\cite{cheng2020learned}) with a quality level of 3), diffusion regeneration~\cite{zhao2025invisible} (60 steps), and center/random crops (0.5/0.7 scale).

\noindent
\textbf{Metrics.}
Detection performance is measured by TPR@1\%FPR. 
FID score~\cite{heusel2017gans} is calculated between the 1,000 watermarked images and corresponding images from the MS-COCO. (FID-1k)

\subsection{Comparison with Baselines}
\label{ssec:exp-baselines}
This section presents a quantitative comparison of our proposed watermarking methodology against state-of-the-art techniques, evaluating resilience, image quality, and computational efficiency.

\noindent
\textbf{Resilience and Image Quality.}
As shown in \Cref{tab:exp_detection}, our performance-centric model (Ours-Perf., APM) achieves a 0.999 verification rate, comparable to leading in-generation methods like SFW and G.Shading. Notably, it achieves this single-shot resilience to regeneration attacks post-hoc, a capability previously considered exclusive to in-generation techniques. Its center-aware design also provides significant defense against cropping attacks, with a success rate approaching 1.0.
Image quality is evaluated using the FID score to ensure a fair comparison with in-generation techniques, which lack reference images. The $\Delta$FID-1k values in \Cref{tab:exp_detection} show that our methodology causes almost no degradation in image quality. Our quality-centric model, Ours-Qual. (PCQ), even slightly improves image quality, with a $\Delta$FID-1k of -0.517. The APM variant shows only a negligible change of +0.042. This performance represents a clear advantage over recent methods such as Tree-Ring (+1.184) and RingID (+2.054), which incur significant quality degradation.

\noindent
\textbf{Computational Efficiency.}
\Cref{tab:exp_computation} compares the computational efficiency of each methodology. End-to-end learning methods require substantial training costs in addition to their inference time. ZoDiac, our most direct comparison, is also a post-hoc latent space method, yet it requires over seven minutes for embedding ($\mathcal{E}$) a single image due to iterative optimization. Semantic-in-generation methods (\textit{Semantic}) require a costly DDIM inversion process for detection ($\mathcal{D}$), averaging four seconds and demanding significant GPU resources per image. In contrast, our method uses only a single, lightweight VAE model, enabling real-time processing for both embedding ($\mathcal{E}$, 0.142s) and detection ($\mathcal{D}$, 0.050s). This demonstrates that high resilience and quality can be achieved simultaneously without heavy computational cost.

\begin{figure}[t]
\centering
\includegraphics[width=\linewidth]{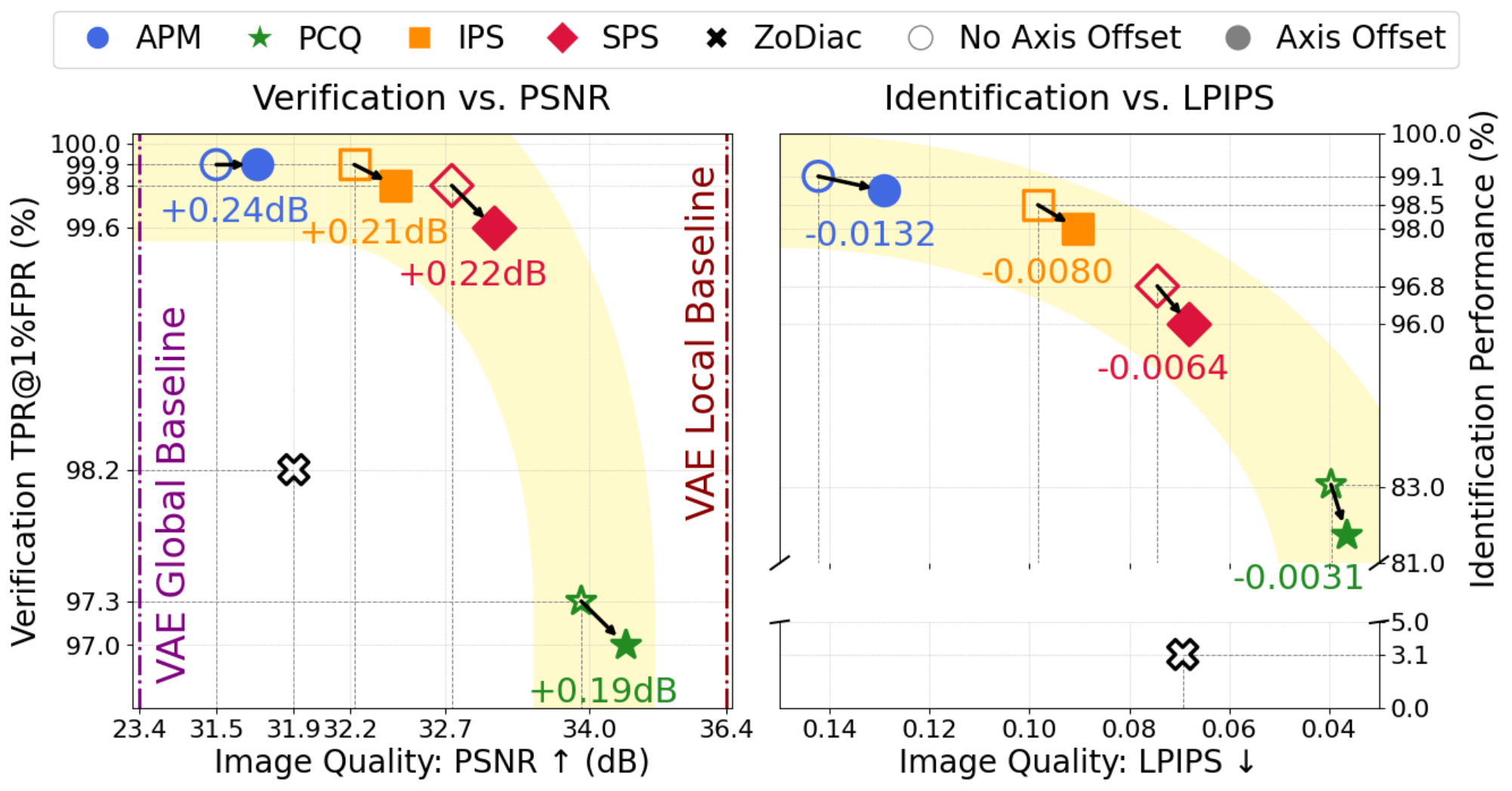}
\caption{Performance vs. image quality trade-off for the four proposed variants (APM, PCQ, IPS, SPS). The comparison includes ZoDiac and illustrates the positive effect of applying axis offsets.}
\label{fig:variants}
\end{figure}

\begin{figure}[t]
\centering
\includegraphics[width=\linewidth]{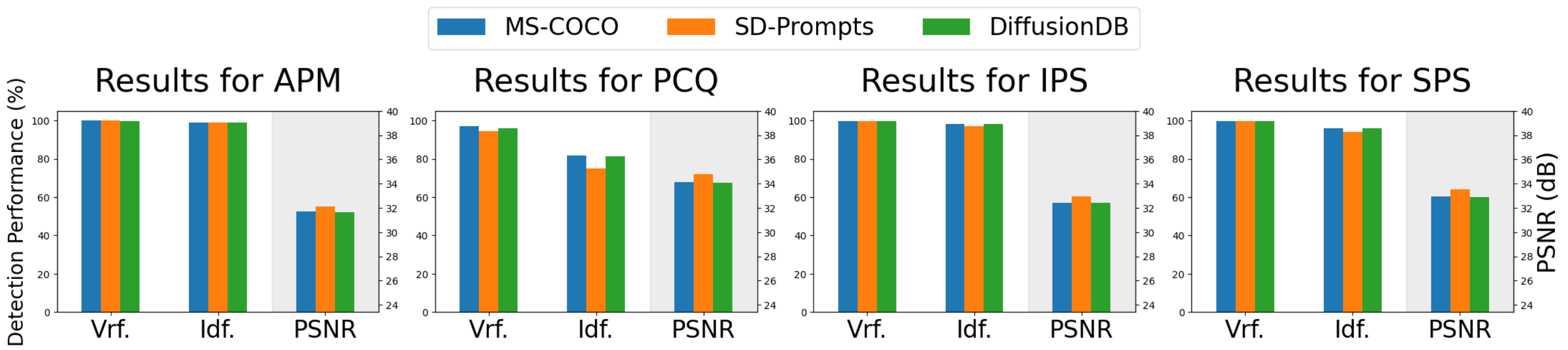}
\caption{Generalization performance. PhaseMark maintains consistent detection rates and image quality across three datasets, demonstrating high stability.}
\label{fig:generalization}
\end{figure}

\subsection{Analysis of Proposed Variants}
\label{ssec:exp-variants}
This section analyzes the performance of the four proposed watermarking variants and validates the effect of a key design component, the \textit{Axis Offset}.

\noindent
\textbf{Trade-off between Detection and Quality.}
\Cref{fig:variants} illustrates the performance-quality relationship of the four proposed variants and compares them with ZoDiac. The left plot shows verification performance versus PSNR, while the right plot shows identification performance versus LPIPS~\cite{zhang2018unreasonable}. The four proposed variants form a distinct trade-off between detection performance and image quality, highlighted by the yellow shaded region in the figure. This trade-off allows users to select an optimal model based on their requirements, ranging from the performance-centric APM to the quality-centric PCQ.
In terms of image quality, our quality-centric variant (PCQ with axis offsets) achieves a PSNR of 34.156. This value is only marginally lower than the VAE's maximum reconstruction fidelity of 36.4193 on our dataset and is substantially higher than the established VAE performance baseline of 23.4 $\pm$ 3.8 on the official benchmark. This demonstrates that the proposed method preserves high fidelity, especially considering the inherent information loss of the underlying VAE.
ZoDiac is included for comparison because it is the only publicly available technique that can embed a watermark into the latent domain as a post-hoc process. ZoDiac performs adequately in the verification task, but its performance collapses to 3.1\% in the identification task, rendering it unusable. This stands in stark contrast to our proposed variants, all of which exhibit strong performance on both tasks.

\noindent
\textbf{Axis offsets.}
All of our final proposed models apply the \textit{Axis Offset} technique. To verify its effect, \Cref{fig:variants} plots each variant with and without the axis offset applied.
As the arrows in \Cref{fig:variants} indicate, applying the axis offset improves image quality (higher PSNR, lower LPIPS) at the cost of a minor drop in detection performance. This is because it avoids modulating high-energy principal frequency axes, minimizing quality degradation. Given the substantial quality gain for a small performance trade-off, we incorporate it into all final models.
This tendency is consistently observed in the results of the subsequent ablation study (\Cref{tab:ablation_ours}).

\noindent
\textbf{Generalization Performance.}
To verify the generalization performance of our methodology, we evaluate it on two additional datasets: SD-Prompts~\cite{gustavo} and DiffusionDB~\cite{wang2022diffusiondb}. The results presented in \Cref{fig:generalization} show that the proposed methods exhibit consistent detection performance and image quality across these diverse datasets, demonstrating high stability.

\setlength{\tabcolsep}{5pt} 
\begin{table}[t]
\centering
\scriptsize
\caption{Ablation study on key parameters for the IPS variant. The results show the impact of channel capacity, axis offsets, and Hermitian symmetry on performance and image quality.}
\begin{tabular}{c ccc|cc|cc}
\arrayrulecolor{black}
\cmidrule[\heavyrulewidth]{2-8}
& $N_{C}$ & Bits & Axis & PSNR $\uparrow$ & LPIPS $\downarrow$ & \textit{Vrf.} & \textit{Idf.} \\
\cmidrule{2-8}
\multirow{10}{*}{\rotatebox{90}{Cut-off Imaginary}} 
& \multirow{2}{*}{1} & \multirow{2}{*}{32} & \ding{55} & 35.271 & 0.0201 & 0.673 & 0.067 \\
& & & $\checkmark$ & \cellcolor{gray!20} 35.356 & \cellcolor{gray!20} 0.0192 & 0.671 & 0.068 \\
\cmidrule{2-8}
& \multirow{2}{*}{2} & \multirow{2}{*}{64} & \ding{55} & 34.875 & 0.0232 & 0.834 & 0.284 \\
& & & $\checkmark$ & \cellcolor{gray!20} 34.988 & \cellcolor{gray!20} 0.0220 & 0.825 & 0.286 \\
\cmidrule{2-8}
& \multirow{2}{*}{3} & \multirow{2}{*}{96} & \ding{55} & 34.299 & 0.0322 & 0.943 & 0.519 \\
& & & $\checkmark$ & \cellcolor{gray!20} 34.463 & \cellcolor{gray!20} 0.0297 & 0.938 & 0.520 \\
\cmidrule{2-8}
& \multirow{2}{*}{4} & \multirow{2}{*}{128} & \ding{55} & 33.808 & 0.0414 & 0.979 & 0.707 \\
& &  & $\checkmark$ & \cellcolor{gray!20} 34.013 & \cellcolor{gray!20} 0.0378 & 0.975 & 0.694 \\
\cmidrule[\heavyrulewidth]{2-8}

\\ [-3mm]

\cmidrule[\heavyrulewidth]{2-8}
\multirow[-5mm]{10}{*}{\rotatebox{90}{Frequency Restored (Ours)}} 
& \multirow{2}{*}{1} & \multirow{2}{*}{32} & \ding{55} & 33.994 & 0.0420 & 0.950 & 0.640 \\
& & & $\checkmark$ & \cellcolor{gray!20} 34.127 & \cellcolor{gray!20} 0.0396 & 0.946 & 0.629 \\
\cmidrule{2-8}
& \multirow{2}{*}{2} & \multirow{2}{*}{64} & \ding{55} & 33.484 & 0.0521 & 0.977 & 0.847 \\
& & & $\checkmark$ & \cellcolor{gray!20} 33.637 & \cellcolor{gray!20} 0.0485 & 0.977 & 0.837 \\
\cmidrule{2-8}
& \multirow{2}{*}{3} & \multirow{2}{*}{96} & \ding{55} & 32.774 & 0.0767 & 0.995 & 0.952 \\
& & & $\checkmark$ & \cellcolor{gray!20} 32.967 & \cellcolor{gray!20} 0.0708 & 0.995 & 0.948 \\
\cmidrule{2-8}
& \multirow{2}{*}{4} & \multirow{2}{*}{128} & \ding{55} & 32.222 & 0.0983 & 0.999 & 0.985 \\
& & & $\checkmark$ & \cellcolor{gray!20} 32.437 & \cellcolor{gray!20} 0.0903 & 0.998 & 0.980 \\
\cmidrule[\heavyrulewidth]{2-8}
\end{tabular}
\label{tab:ablation_ours}
\end{table}
\subsection{Ablation Study}
\label{ssec:exp-ablations}
This section presents the results of an ablation study on the key parameters of our methodology. For brevity, we summarize the results for IPS, the variant with the most balanced performance, as a representative case in \Cref{tab:ablation_ours}. All other variants exhibit consistent trends in response to parameter changes.

\noindent
\textbf{Channel Capacity.}
Our method embeds a 32-bit message per channel in the latent vector. We analyze the impact of increasing the number of channels ($N_C$), which directly corresponds to the total bit capacity. \Cref{tab:ablation_ours} shows that as $N_C$ increases from 1 to 4, the bit capacity increases from 32 to 128. As expected, this results in a trade-off: detection performance (\textit{Vrf.}, \textit{Idf}.) improves while image quality (PSNR, LPIPS) degrades. A shorter watermark length requires a higher detection threshold to control the false positive rate, which can weaken detection power. We thus determine that the 128-bit setting ($N_C = 4$) is the most suitable, as it ensures stable detection performance.

\noindent
\textbf{Hermitian Symmetry for Real-Valued Output.}
We validate enforcing Hermitian symmetry against naively discarding the imaginary part (Cut-off Imaginary). As shown in \Cref{tab:ablation_ours}, Hermitian symmetry is crucial for detection, boosting Verification/Identification performance from 0.975/0.694 to 0.998/0.980. The naive approach dilutes the embedded signal, which degrades detection. The paradoxically higher image quality of the naive method simply indicates that the watermark is not embedded correctly. Thus, enforcing symmetry is essential for reliable watermarking.

\section{Conclusion}
\label{sec:conclusion}
In this work, we introduced PhaseMark, a novel post-hoc watermarking framework that achieves state-of-the-art resilience and efficiency without iterative optimization. By directly modulating phase in the latent frequency domain, PhaseMark overcomes the prohibitive computational costs of prior methods, enabling real-time performance while preserving image quality. Our analysis of four variants provides a clear performance-quality trade-off for practical applications. For enhanced security, the watermark payload can be encrypted using cryptographic algorithms like ChaCha20 to ensure confidentiality. PhaseMark presents an efficient new paradigm, paving the way for the large-scale deployment of robust watermarking in the generative AI era.


\vfill\pagebreak

\bibliographystyle{IEEEbib}
\bibliography{strings, refs}

\end{document}